\documentclass[a4paper,twocolumn]{article}

\usepackage[T1]{fontenc}
\usepackage{lmodern}
\usepackage[numbers]{natbib}
\usepackage{amsfonts}
\usepackage{amsmath}
\usepackage{booktabs}
\usepackage{graphicx}
\usepackage{subcaption}
\usepackage[hidelinks]{hyperref}
\usepackage{flushend}

\title{Deep Reinforcement Learning for Multi-UAV Data Harvesting Under Variable Team Sizes and Agent Loss}
\author{%
  \begin{tabular}{@{}ccc@{}}
    Thiago de Souza Lamenza\thanks{Corresponding author.} &
    Bruno José Olivieri de Souza & Markus Endler \\
    \small\href{mailto:tlamenza@inf.puc-rio.br}{tlamenza@inf.puc-rio.br} &
    \small\href{mailto:bolivieri@inf.puc-rio.br}{bolivieri@inf.puc-rio.br} &
    \small\href{mailto:endler@inf.puc-rio.br}{endler@inf.puc-rio.br} \\
    \footnotesize\href{https://orcid.org/0009-0001-1510-6735}{ORCID: 0009-0001-1510-6735} &
    \footnotesize\href{https://orcid.org/0000-0002-1707-7755}{ORCID: 0000-0002-1707-7755} &
    \footnotesize\href{https://orcid.org/0000-0002-8007-9817}{ORCID: 0000-0002-8007-9817}
  \end{tabular}\\[0.5em]
  \small Pontifícia Universidade Católica do Rio de Janeiro, Rio de Janeiro, RJ, Brazil
}
\date{}

\begin{document}
\maketitle

\begin{abstract}
Collecting data from wireless sensors can be challenging when limited communication infrastructure prevents them from offloading their measurements. In such settings, unmanned aerial vehicles (UAVs) can exploit their mobility to harvest data locally, while coordinated UAV teams can provide efficient and redundant coverage of remote areas. Multi-agent reinforcement learning (MARL) enables such teams to learn decentralized coordination policies. However, training separate policies for different team and sensor populations is costly, and fixed-population policies may be vulnerable to agent failures. This work proposes F-MADDPG, a flexible variant of Multi-Agent Deep Deterministic Policy Gradient (MADDPG) that supports data harvesting with varying numbers of wireless sensors and UAVs. The framework also accommodates unexpected catastrophic failures that permanently remove agents during a mission. We compare F-MADDPG with a MADDPG baseline that uses multilayer perceptrons (MLPs) and evaluate training strategies based on team-size sampling and exposure to agent terminations. The results show that F-MADDPG learns across the evaluated population sizes, completes missions faster than the baseline on average, benefits from inverse-cardinality team-size sampling compared with uniform sampling, remains effective after permanent agent losses, and gains no consistent benefit from exposure to failures during training.
\end{abstract}

{\small\noindent\textbf{Keywords:} Distributed systems, data harvesting, reinforcement learning, fault tolerance\par}

\section{Introduction}
Unmanned aerial vehicles (UAVs) provide a mobile platform for tasks performed over large or difficult-to-access areas \cite{cavalcantiRescueFromTheAir2026,reimerFromAirToGround2025,samirAgeInformationAware2020,messaoudiSurveyUAVbasedData2023}. When deployed as a team, they can divide spatially distributed work and continue operating without relying on permanent infrastructure. One relevant application is data harvesting from wireless sensor networks: UAVs travel between ground sensors and collect their measurements through short-range air-to-ground communication \cite{benghorbelEnergyEfficientData2018,bouhamedUAVAssistedDataCollection2020,desouzaCollectingSensorData2023}. Using a UAV swarm can shorten a mission, but they must coordinate their actions to complete the task efficiently, avoiding redundant sensor visits and distributing the remaining workload.

Designing coordination rules by hand becomes difficult as the number of UAVs and sensors increases and as their spatial configuration changes. Multi-agent reinforcement learning (MARL) offers an alternative in which coordinated behavior is learned through interaction with a simulated environment. Deep reinforcement-learning methods have consequently been applied to multi-UAV data collection, jointly addressing decisions such as trajectory planning, sensor selection, and task allocation \cite{khodaparastDeepReinforcementLearning2021,wangCooperativeDataCollection2023,wuSemanticsAwareMultiUAVCooperation2023,leeMultiAgentDeepReinforcement2024}.

The flexibility obtained through learning comes at the cost of an expensive training process. Deep reinforcement-learning policies may require millions of environment interactions, and a policy trained for one scenario configuration may perform poorly when deployed under substantially different conditions. Training a separate model for every possible team size in a multi-agent problem would multiply this cost and require the system to maintain and select among several population-specific policies. A single policy capable of operating across team sizes can instead reuse training across configurations and makes the learned controller easier to deploy.

Population variability is also a fault tolerance concern: UAVs may become unavailable because of energy depletion, communication problems, adverse weather, or equipment failure. Consequently, solutions that can adapt to dynamic agent populations are better equipped to handle such conditions. A permanent failure during an episode abruptly changes the number of participating agents and requires the survivors to redistribute unfinished work. However, conventional MARL implementations commonly represent inputs using fixed-dimensional vectors tied to a predefined number of agents. Merely filling an absent agent's slot with a fixed value does not explicitly distinguish it from valid input and does not eliminate the relationship of the network architecture with a maximum population.

In our previous work \cite{desouzalamenzaDeepReinforcementLearning2026} we proposed two algorithms that addressed population variability. The first was successful in producing a model that works across a range of sensor populations. The second attempted the same for UAV populations, but ultimately failed to achieve consistent performance across team sizes.

We now propose F-MADDPG, a flexible variant of Multi-Agent Deep Deterministic Policy Gradient (MADDPG) \cite{loweMultiAgentActorCriticMixed2017} for cooperative UAV data harvesting under variable team sizes and permanent agent losses. F-MADDPG integrates a Flexible Encoder into a shared decentralized actor and centralized critic. The encoder uses MLP heads for fixed-size fields and masked attention heads for variable-cardinality fields, producing a fixed-size representation for each field. Explicit validity masks exclude absent agents and unavailable sensors, allowing one policy to operate across the evaluated populations.

We evaluate F-MADDPG in a simulated data-harvesting task with varying numbers of UAVs and sensors and with stochastic, permanent UAV losses. Our experiments compare F-MADDPG with a MADDPG baseline that uses multilayer perceptrons (MLPs) in terms of learning efficiency and final task performance. We also formulate an inverse-cardinality strategy for sampling initial team sizes during training and compare it with uniform sampling to evaluate its effect on performance across team sizes. Finally, we compare policies trained with different failure probabilities to investigate whether explicit failure exposure provides benefits beyond training across initial team sizes. F-MADDPG learns effective policies across the evaluated configurations, completes missions faster than the MLP baseline on average, and benefits from inverse-cardinality team-size sampling. For each policy, performance remains similar across the evaluated failure probabilities. Exposure to failures during training does not consistently improve performance.

\section{Related Work}
Multi-agent reinforcement learning has been applied to UAV data harvesting to coordinate movement and collection decisions under objectives such as energy efficiency and information freshness \cite{khodaparastDeepReinforcementLearning2021,wangCooperativeDataCollection2023,wuSemanticsAwareMultiUAVCooperation2023,leeMultiAgentDeepReinforcement2024}. These studies address important operational constraints, including limited energy and communication range. A further question concerns coordination across changing team sizes: the number of UAVs may differ between missions, or the active team may change during a mission. Permanent UAV loss is one important instance of the latter, requiring the surviving agents to complete work that would otherwise have been performed by the missing teammate.

Ramezani et al. \cite{ramezaniFaultTolerantMultiAgentReinforcement2024} enable surviving UAVs to continue a coverage mission after permanent losses by switching among population-specific configurations. This accommodates within-mission changes but requires training, maintaining, and selecting among several population-specific policies. Our prior work \cite{desouzalamenzaDeepReinforcementLearning2026} explored training models with a fixed team size and deploying them in scenarios with different UAV counts. In our experiments, this strategy was effective when the deployment population was at least as large as the training population, although transfer to larger teams incurred a completion-time penalty. Authors in \cite{bayerleinMultiUAVPathPlanning2021,seongMultiUAVTrajectoryOptimizer2023} represent a rectangular collection area as a discrete grid, absorbing population variability into a task-specific spatial representation. They explore varying populations between episodes but not within episodes. The spatial representation imposes restrictions on the shape of the environment, limits the spatial resolution with which UAV positions are represented, and requires a task which can be represented in a spatially structured manner. 

Looking beyond data harvesting, the cooperative MARL literature offers several strategies for accommodating varying agent populations. Some studies test transfer between fixed training and deployment populations without changing the underlying architecture \cite{gurestiEvaluatingGeneralizationTransfer2021}. Such transfer is useful when local observations remain comparable, but it does not address recovery from agent losses. Other methods construct bounded local views. K-nearest MARL \cite{khorasganiKnearestMultiagentDeep2021} forms fixed-size neighborhoods within a larger variable population. Tong et al. \cite{tongEnergyawareFaulttolerantDeep2026} use a spatial representation, similar to those discussed above, for cooperative patrolling, evaluating continued operation after catastrophic agent failures and subsequent additions. Their shared actor incorporates teammate locations into a fixed-size map, while their centralized critic retains explicit team information padded to a maximum training population.

In the broader reinforcement learning literature, some authors explore neural architectures that explicitly accommodate variations in agent population. Long et al. \cite{longEvolutionaryPopulationCurriculum2020} propose a population-invariant MADDPG architecture with two attention-based aggregation stages in the critic. The first aggregates observation entities to form each agent's observation--action embedding, while the second aggregates agent embeddings into a fixed-size representation for the centralized critic. The actor also uses attention over observation entities. F-MADDPG instead centralizes corresponding observation and action fields before field-specific encoding and shares parameters across agents rather than maintaining independently parameterized agents. Long et al.'s main objective is to scale MARL training to larger populations through Evolutionary Population Curriculum (EPC). Across stages of increasing population size, EPC combines candidate agent sets, fine-tunes them, and selects those that adapt best to the enlarged population. The reported experiments show better performance at larger populations than training from scratch, and an extrapolation test deploys cloned policies at twice the largest training population without further training. Some of the experiment environments also include game-driven agent deaths, although operational fault tolerance is not the study's focus. Our work investigates a shared policy across team sizes and continued UAV data harvesting after permanent agent losses.

Self-attention is also used in the MA-POCA algorithm \cite{cohenUseMisuseAbsorbing2021}, which was designed for cooperative environments in which agents can spawn or terminate during an episode. Its centralized value and counterfactual-baseline networks aggregate information from currently active agents, avoiding explicit absorbing-state inputs for terminated agents. The authors also address posthumous credit assignment, which allows an agent's earlier decisions to receive credit from team rewards obtained after its termination. Our work also investigates attention-based aggregation of variable-cardinality inputs, applying it to both the actor and critic. Our experiments did not reveal evidence that posthumous credit assignment was a practical issue in the evaluated data-harvesting setting.

Finally, Alarcon Granadeno et al. \cite{alarcongranadenoMultisourcePlumeTracing2025} use recurrent neural networks to process observations from a variable number of teammates in a multi-UAV plume-tracing task. Their Action-Specific Double Deep Recurrent Q-Network (ADDRQN) architecture combines separate embeddings of the acting UAV's current observation and previous action with a fixed-size teammate representation produced by a gated recurrent unit (GRU). These embeddings are passed to a long short-term memory (LSTM) network, followed by an output network estimating Q-values for seven discrete actions. All UAVs share the same network parameters. Whereas the teammate GRU accommodates population variability, the temporal LSTM maintains memory under partial observability. Our method instead uses attention-based entity encoding within a continuous-action actor--critic formulation. Their described training randomizes initial team sizes without explicit within-mission failure injection, and evaluation includes permanent UAV dropout and communication blackouts. Our experiments focus on permanent losses and additionally compare different training-time failure-exposure conditions.

These studies also address how different team sizes are represented during learning. Alarcon Granadeno et al. partition replay experience into team-size buckets and uniformly select a bucket before sampling training sequences, while Tong et al. \cite{tongEnergyawareFaulttolerantDeep2026} allocate additional parallel training episodes to single-agent teams. Our inverse-cardinality strategy instead changes the initial-team sampling distribution to compensate for larger teams contributing more agent-level experience per environment step. This concerns experience collection, whereas ADDRQN's buckets act during replay sampling. Episode duration and agent losses also affect the amount of collected experience, so initial-team sampling alone does not guarantee a balanced final distribution. In Section~\ref{sec:team-sampling}, we compare inverse-cardinality sampling with uniform initial-team sampling and an adaptation of ADDRQN's bucketed replay strategy within F-MADDPG. These experiments do not compare the complete F-MADDPG and ADDRQN algorithms.

Overall, these studies highlight two related concerns: training policies that generalize to variable agent populations and adapting to within-episode changes in active agent count. Our evaluation brings these concerns together in UAV data harvesting by examining the training efficiency of our proposed actor–critic representation, the distribution of initial team sizes during training, and exposure to permanent within-mission losses.

\section{Algorithm}
\subsection{Background}

A cooperative multi-agent reinforcement-learning task can be modeled as a partially observable Markov game. The objective is to maximize the expected discounted team return

\begin{equation}
J(\pi)=\mathbb{E}_{\pi}\!\left[\sum_{t=0}^{T-1}\gamma^t r_t\right],
\end{equation}

where $\pi$ denotes the policy, $r_t$ is the shared team reward at time $t$, $\gamma\in[0,1)$ is the discount factor, and $T$ is the episode horizon \cite{suttonReinforcementLearningIntroduction2018}.

We use MADDPG \cite{loweMultiAgentActorCriticMixed2017} with centralized training and decentralized execution. Each UAV uses a shared actor $\mu_\phi$ to compute a continuous action based on its local observation, while a centralized critic $Q_\theta$ evaluates the active team's joint observations and actions during training. Parameter sharing \cite{chuParameterSharingDeep2017} exploits the homogeneity of the UAVs and allows active agents to execute the same policy independently.

Let $\mathcal{I}^{\max}=\{1,\ldots,N_a^{\max}\}$ denote the available agent slots and $\mathcal{I}_t\subseteq\mathcal{I}^{\max}$ the active agents at time $t$. A failure is applied after the affected UAV acts in transition $t$. With $D_t$ indicating completion or truncation of the team episode, we define
\begin{equation}
\begin{aligned}
m_t^j &= \mathbb{I}\!\left[j\in\mathcal{I}_t\right],\\
e_t^j &= m_t^j\left(1-m_{t+1}^j\right),\\
d_t^j &= D_t\lor e_t^j,\\
y_t^j &= r_t+\gamma(1-d_t^j)
\bar{Q}\!\left(\mathbf{o}_{t+1},
\bar{\boldsymbol{\mu}}(\mathbf{o}_{t+1}),
\mathbf{m}_{t+1}\right).
\end{aligned}
\end{equation}
Here, $\mathbb{I}$ is the indicator function, which returns 1 if its condition is true and 0 otherwise. Thus, $m_t^j$ indicates whether slot $j$ is active at time $t$, and $e_t^j$ indicates whether the agent terminates during the transition. Slots absent at episode reset represent padding rather than failures, but both absent and failed slots are excluded from centralized inputs and loss reductions.

The centralized critic produces one value for the team, and $y_t^j$ is its TD(0) training target for agent $j$. At time $t+1$, $\mathbf{o}_{t+1}$ denotes the joint observations. The joint action $\bar{\boldsymbol{\mu}}(\mathbf{o}_{t+1})$ collects the actions produced by applying the shared target actor $\bar{\mu}$ to each agent's local observation. The target critic $\bar{Q}$ evaluates the joint observations and actions. The terminal indicator $d_t^j$ prevents a failed UAV from bootstrapping beyond its final reward, while survivors bootstrap using the reduced team if the episode continues. Partial agent loss therefore does not terminate the team episode, allowing the remaining UAVs to redistribute unfinished work.

Actor and critic losses are computed using masked mean reduction:

\begin{equation}
L=\frac{\sum_{t,j}m_t^j\,\ell_t^j}{\sum_{t,j}m_t^j},
\end{equation}

where $\ell_t^j$ denotes the corresponding unreduced per-agent loss, and the sums run over the sampled transitions and agent slots in the minibatch. An agent's final active transition remains included because $m_t^j=1$; subsequent entries have $m_t^j=0$ and contribute no gradient.

\subsection{Flexible Encoder} \label{sec:flex-enc}

Conventional deep reinforcement-learning networks typically flatten observations into fixed-dimensional vectors. Padding variable entity sets to a maximum size provides regular tensors, but an MLP cannot inherently distinguish padding from valid values or model interactions among the entities.

\begin{figure*}[t]
\centering
\includegraphics[width=0.9\textwidth]{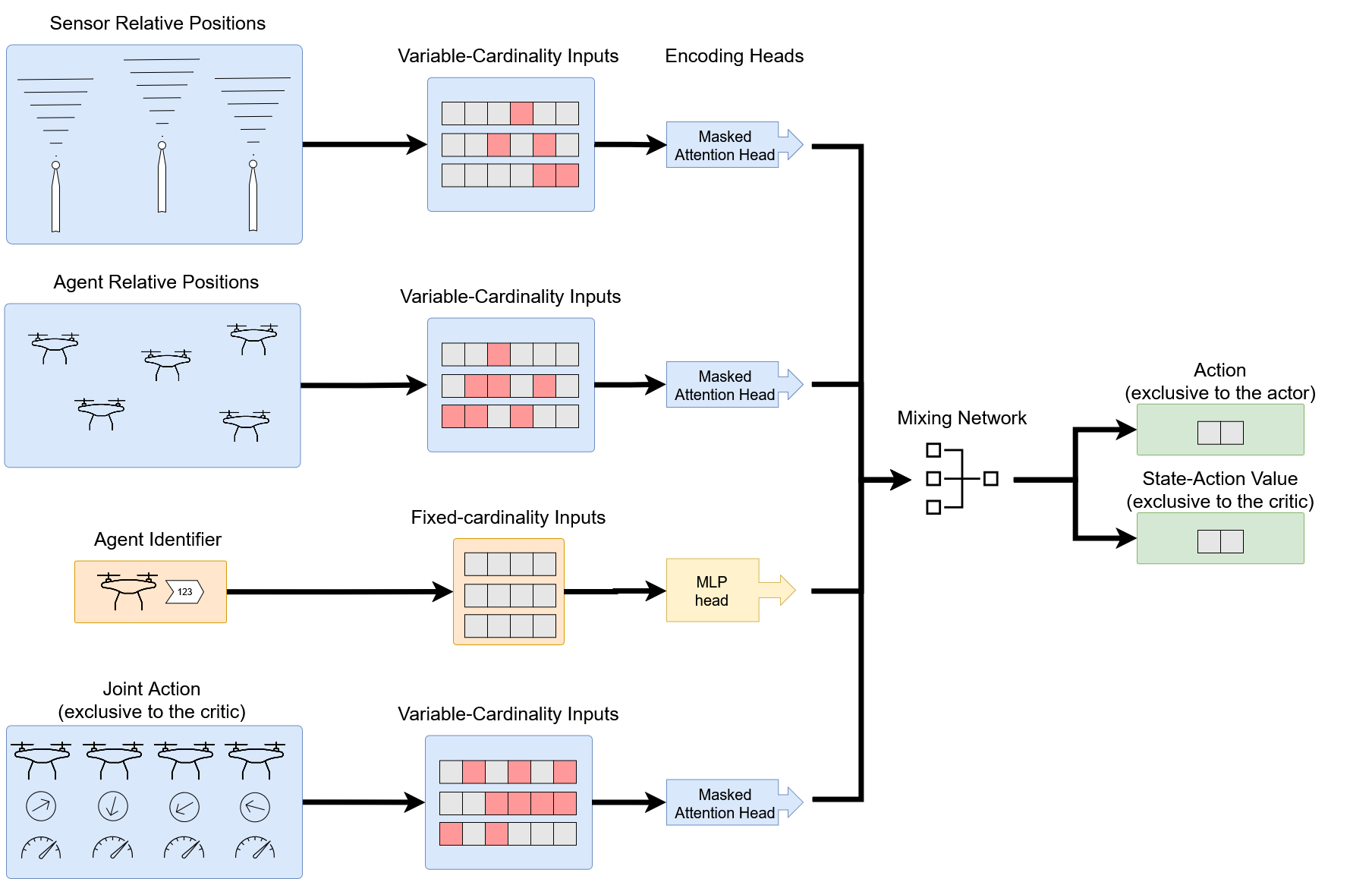}
\caption{Diagram of the Flexible Encoder architecture and how it is applied to the data-harvesting task. Inputs with variable-cardinality such as sensor and teammate positions are processed by masked attention heads while fixed-size inputs such as UAV identifiers are processed by MLP heads. The resulting representations are concatenated and mixed to produce a fixed-size output.}
\label{fig:flex_encoder}
\end{figure*}

The \emph{Flexible Encoder} instead decomposes an observation $o$ into $H$ semantically distinct fields. As shown in Figure~\ref{fig:flex_encoder}, a field-specific head $f_h$ maps the input $x_h$ to a fixed-size representation. For variable-cardinality fields, the head also receives a validity mask $b_h$ that identifies the valid entries in $x_h$. The resulting representations are concatenated and mixed:

\begin{equation}
\begin{aligned}
z_h&=f_h(x_h,b_h)\in\mathbb{R}^{n_h},\\
E(o)&=f_{\mathrm{mix}}\!\left(z_1\mathbin{|}\cdots\mathbin{|}z_H\right)
\in\mathbb{R}^{n_{\mathrm{mix}}},
\end{aligned}
\end{equation}

where $\mathbin{|}$ denotes concatenation and $b_h$ is omitted for fixed-size fields. Consequently, the encoder output dimension is independent of the number of valid entities. The encoder does not prescribe a reinforcement learning update rule. Its output can be used wherever a conventional neural network would otherwise consume a flattened observation.

The MLP heads encode fixed-size fields such as scalars or single-entity feature vectors. Masked attention heads encode variable-cardinality fields by projecting each element to the Transformer embedding dimension, applying self-attention using $b_h$ to exclude padding, and performing masked mean pooling. If $b_h$ contains no valid elements, the head returns the zero vector. Padding therefore affects neither attention nor pooling. Entity identity embeddings may be added before attention when entity roles must be retained. The pooled representation is independent of the ordering of entities.

In the data-harvesting implementation, the actor uses an MLP head for the UAV identifier and masked attention heads for the relative positions of sensors and other UAVs. Their validity masks, described in Section~\ref{sec:environment-integration}, exclude collected or nonexistent sensors and inactive or nonexistent agents. The following subsection specifies the separate actor and critic encoders used by F-MADDPG.

\subsection{F-MADDPG}

F-MADDPG replaces the conventional actor and critic feature networks of MADDPG with separate Flexible Encoders $E^\mu_\phi$ and $E^Q_\theta$. The shared actor processes each agent's local observation, whereas the critic produces a scalar value from centralized observations and actions.

The actor encoder produces a vector with the same dimensionality as the action space. Its output is normalized element-wise to $[0,1]$:

\begin{equation}
a_t^j=\mu_\phi(o_t^j)=\rho\!\left(E^\mu_\phi(o_t^j)\right),
\quad
\rho(v)=\frac{\tanh(v)+1}{2}.
\end{equation}

To construct the critic's input, a centralization operator $\mathcal{C}$ concatenates corresponding fields across agent slots while combining each field mask $b^j_{h,t}$ with the activity mask $m_t^j$. For a variable-cardinality observation field, the centralization operation and resulting critic evaluation are given by

\begin{equation}
\begin{gathered}
\mathcal{C}_h(\mathbf{o}_t,\mathbf{m}_t)
=\left(
\operatorname*{concat}_{j=1}^{N_a^{\max}}x^j_{h,t},
\operatorname*{concat}_{j=1}^{N_a^{\max}}
\left(m_t^j\land b^j_{h,t}\right)
\right),\\
Q_\theta(\mathbf{o}_t,\mathbf{a}_t,\mathbf{m}_t)
=E^Q_\theta\!\left(
\mathcal{C}(\mathbf{o}_t,\mathbf{m}_t),
(\mathbf{a}_t,\mathbf{m}_t)
\right).
\end{gathered}
\end{equation}

Here, $\mathbf{o}_t$ and $\mathbf{a}_t$ contain the observations and actions of all available slots. Concatenating a fixed-size observation field across agents yields a variable-cardinality field whose valid entries correspond to active agents. The joint action is an additional variable-cardinality field masked by $\mathbf{m}_t$. Each centralized element receives a learned embedding identifying its owning agent. Padding and information from inactive agents are excluded while the critic can still distinguish the sources of valid elements.

\subsection{Data-Harvesting Task}
\label{sec:environment-integration}

The task builds on our previous study \cite{desouzalamenzaDeepReinforcementLearning2026}, retaining proximity-based sensor collection, shared team information, and the cooperative objective of completing collection efficiently. Here, we investigate variable active populations and permanent within-mission UAV losses.

The task contains a variable-size team of UAVs and stationary ground sensors at known positions. Each sensor is collected once, and the numbers and positions of UAVs and sensors are sampled at each episode reset. UAVs are assumed to share their positions and sensor-collection status through reliable team communication. The observation of active UAV $j$ is
\begin{equation} \label{eq:observation}
o_t^j =
\left\langle
\mathcal{P}_{s,t}^j,\mathbf{b}_{s,t}^j,
\mathcal{P}_{a,t}^j,\mathbf{b}_{a,t}^j,
\iota_t^j
\right\rangle ,
\end{equation}
where $\mathcal{P}_{s,t}^j$ contains the normalized relative positions of the $N_s^*$ closest uncollected sensors and $\mathcal{P}_{a,t}^j$ those of the $N_a^*$ closest other active UAVs. We set $N_s^*=12$ and use $N_a^*=7$ unless otherwise specified. The Boolean masks $\mathbf{b}_{s,t}^j$ and $\mathbf{b}_{a,t}^j$ distinguish valid entities from padding, while $\iota_t^j$ is the UAV's identifier. Collected sensors and inactive UAVs are excluded. Consequently, an agent failure immediately removes that UAV from subsequent observations.

Each actor produces a normalized two-dimensional action
\begin{equation}
\begin{aligned}
a_t^j&=\langle a_{\theta,t}^j,a_{v,t}^j\rangle\in[0,1]^2,\\
\theta_t^j&=2\pi a_{\theta,t}^j,\\
v_t^j&=v_{\max}a_{v,t}^j,
\end{aligned}
\end{equation}
which determines the UAV's movement direction $\theta^j_t$ and speed $v^j_t$. UAVs fly at a fixed altitude and, under a simplified motion model, may change direction and speed instantaneously. Agents must remain within the scenario boundaries. Data collection requires no explicit action: any uncollected sensor within the 20~m horizontal communication range transmits its data and is collected automatically.

All UAVs receive the same team reward. Let $C_t$ and $C_{t+1}$ be the cumulative numbers of sensors collected before and after transition $t$, respectively, and $N_s$ the number of sensors present in the episode. The reward for transition $t$ is
\begin{equation}
r_t =
\begin{cases}
10(C_{t+1}-C_t), & C_{t+1}>C_t,\\[2mm]
-\dfrac{N_s-C_{t+1}}{N_s}, & \text{otherwise}.
\end{cases}
\end{equation}
This rewards newly collected data while penalizing delays in completing the remaining workload. UAVs that fail after acting receive the reward from their final transition; thereafter, their observations, actions, and rewards are masked, and the surviving team continues the mission.

\section{Experimental Setup}
All models evaluated in the following section were trained for 25 million time steps on the same machine, equipped with an RTX 3090 GPU. All training parameters, trained models, simulation code, and algorithms are open source and publicly available. The code and trained policies are available on Zenodo \cite{lamenzaReplicationPackageDeep2026}.

Each condition uses one training run and its resulting policy, with repeated policy evaluations. Reported confidence intervals reflect evaluation-episode variability, not variability across independent training runs.

Unless otherwise specified, we evaluated each policy on 2,000 episodes per evaluation setting. By default, experiments consider team sizes in the range $\left[N_a^{\min}, N_a^{\max}\right]$, with $N_a^{\min}=1$ and $N_a^{\max}=8$, and sensor counts between $N^{\min}_s=1$ and $N^{\max}_s=36$. These variations cover a wide range of UAV and sensor densities. The model must generalize across them to ensure reliable, high-quality task completion.

Completion time is the elapsed simulated time, in seconds, at the end of the step in which the final sensor is collected. Unsuccessful episodes are assigned the 500~s episode horizon.

\section{Results}
\subsection{Training Efficiency} \label{sec:training-efficiency}
Training a deep reinforcement learning model can be highly resource intensive, depending on the complexity of the environment and model architecture \cite{openaiDota2Large2019,vinyalsGrandmasterLevelStarCraft2019}. Consequently, improving sample efficiency can substantially reduce training costs. The approach introduced in Section~\ref{sec:flex-enc} uses specialized heads for each observation field, potentially increasing training efficiency.

To evaluate training efficiency, we compare F-MADDPG with a MADDPG baseline that uses an MLP to process all observations. Because this baseline cannot mask inactive observation entries or agents, these entries are padded with a fixed value of $-1$. Both models use inverse-cardinality team-size sampling and are trained without agent failures. Both of these details will be explored in sections \ref{sec:team-sampling} and \ref{sec:failure-exposure}, respectively.

\begin{figure}[t]
\centering
\includegraphics[width=\columnwidth]{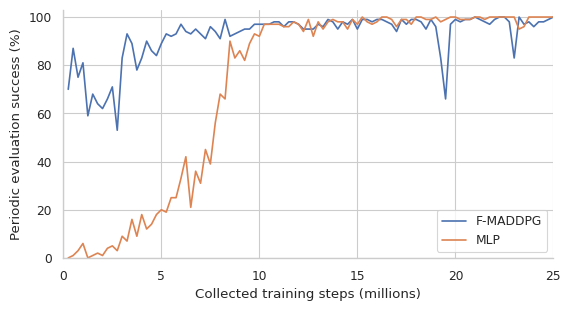}
\caption{Periodic evaluation success rate of F-MADDPG and the MLP baseline over 25 million training steps.}
\label{fig:training-metrics}
\end{figure}

Figure~\ref{fig:training-metrics} presents the success rates of F-MADDPG and the MLP baseline during training. An episode is considered successful if all sensors are visited within the allotted time. The models were periodically evaluated by pausing parameter updates and running 100 episodes with uniformly sampled agent counts, sensor counts, and sensor locations. 

Both F-MADDPG and the baseline achieved near-100\% success rates by the end of the 25 million training steps. The first sequence of three consecutive evaluations with at least 90\% success began at 5.3 million training steps for F-MADDPG and 9.8 million for the baseline. These results suggest that F-MADDPG learned to process the variable-cardinality observation fields more quickly than the baseline.

Success rate alone does not fully characterize task performance because it captures only binary task completion and does not reflect completion quality. Moreover, metrics aggregated over uniformly sampled episodes may conceal weaknesses in specific environment configurations. 

\begin{figure*}[t]
\centering
\begin{subfigure}[t]{0.48\textwidth}
\centering
\includegraphics[width=\linewidth]{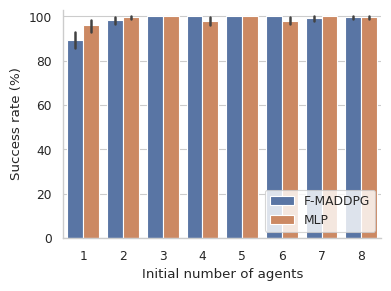}
\caption{Success rate.}
\label{fig:final-encoder-success}
\end{subfigure}
\hfill
\begin{subfigure}[t]{0.48\textwidth}
\centering
\includegraphics[width=\linewidth]{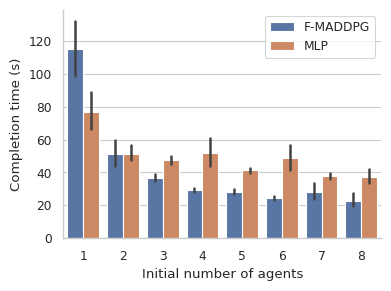}
\caption{Completion time.}
\label{fig:final-encoder-completion}
\end{subfigure}
\caption{Performance of the final F-MADDPG and MLP baseline policies, stratified by initial number of agents. Error bars show 95\% confidence intervals.}
\label{fig:final-encoder-results}
\end{figure*}

Figure~\ref{fig:final-encoder-results} shows the success rate and completion time, stratified by the initial number of agents. Both algorithms achieved high success rates. The largest difference occurred in single-agent episodes, for which F-MADDPG's success rate was 6.5 percentage points lower than that of the baseline. By contrast,  F-MADDPG reduced the mean completion time by 7.2~s across all team sizes.

\subsection{Team Size Sampling}
\label{sec:team-sampling}

Our previous work found that policies trained with larger UAV teams would perform poorly when deployed with fewer agents \cite{desouzalamenzaDeepReinforcementLearning2026}, but even policies trained with smaller teams still suffer from task performance degradation. Here, we expose models to all initial team sizes during training and investigate how the per-episode team size sampling distribution affects performance. 

Because F-MADDPG uses parameter sharing and all agents are homogeneous, each step involving $N_a$ active agents contributes $N_a$ experiences. Uniform initial-team sampling therefore yields more agent-level experience per step from larger teams, potentially leaving smaller teams underrepresented during learning.

Alarcon Granadeno et al. \cite{alarcongranadenoMultisourcePlumeTracing2025} propose ADDRQN, which addresses a similar issue by partitioning replay experience into team-size buckets and uniformly selecting a bucket before sampling training sequences. Our approach instead modifies the initial-team sampling distribution to increase exposure to smaller teams during experience collection.

We sample the initial agent count $N_a$ with a probability inversely proportional to the team size to increase exposure to smaller teams during experience collection:
\begin{equation}
P(N_a) = \frac{1}{N_aZ},
\qquad
Z = \sum_{k=N_a^{\min}}^{N_a^{\max}} \frac{1}{k}.
\end{equation}

Assuming equal average episode duration across team sizes and ignoring within-episode agent losses, the probability $P^e(N_a)$ that a uniformly sampled replay experience was generated by a team of size $N_a$ is
\begin{equation}
P^e(N_a)
= \frac{N_a P(N_a)}{\sum_{k=N_a^{\min}}^{N_a^{\max}} kP(k)}
= \frac{1}{N_a^{\max}-N_a^{\min}+1},
\end{equation}
which is constant across team sizes under these simplifying assumptions. In practice, episode duration and agent losses can change the collected distribution, so inverse-cardinality sampling does not guarantee a balanced final replay memory.

\begin{figure}[t]
\centering
\includegraphics[width=\columnwidth]{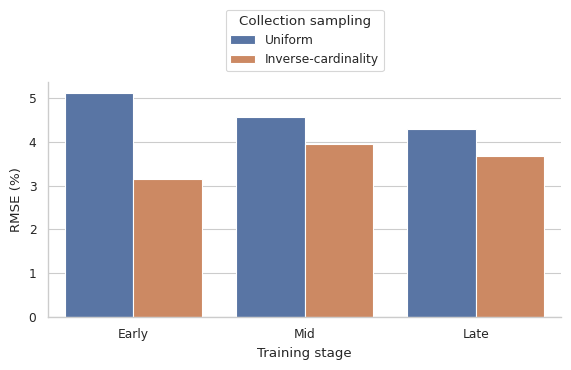}
\caption{Team-size imbalance in collected experience at approximately 4, 12, and 21 million training steps. Bars show root mean squared error (RMSE) relative to an equal representation target (12.5\% per team size), in percentage points.}
\label{fig:sampling-error}
\end{figure}

\begin{figure}[t]
\centering
\includegraphics[width=\columnwidth]{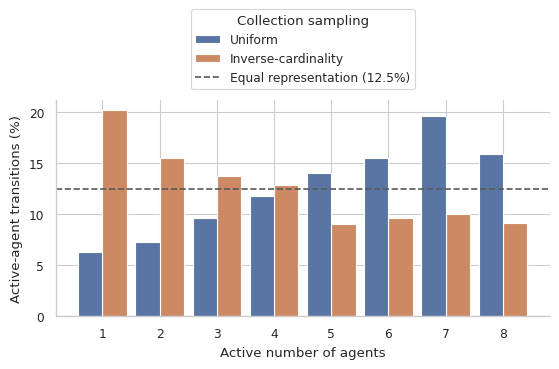}
\caption{Share of collected transitions by active team size in rollouts from the late checkpoint (approximately 21 million training steps). The dashed line indicates equal representation (12.5\% per team size).}
\label{fig:sampling-distribution}
\end{figure}

To assess how initial team-size sampling affects the balance of collected experience, we compared policies trained with uniform and inverse-cardinality sampling. We chose three milestones to represent early, mid, and late training at approximately 4, 12, and 21 million training steps, respectively. For each policy at each milestone, we collected a fresh rollout of 50,000 environment steps under its team-size sampling strategy. Each step contributed one transition per active agent to the count for that step's team size. We computed the root mean squared error (RMSE) of the resulting percentages relative to an equal share of 12.5\% per team size. Figure~\ref{fig:sampling-error} shows that inverse-cardinality sampling yielded lower RMSE at all three milestones. The reduction relative to uniform sampling was largest at the early milestone (38\%).

Figure~\ref{fig:sampling-distribution} shows how the strategies differ at the late checkpoint. With inverse-cardinality sampling, one- and two-agent teams each contributed more than the target share of 12.5\%, while teams of five to eight agents each contributed less. The earlier checkpoints showed a similar shift toward smaller teams. One possible explanation is that episode durations are more similar across team sizes early in training. If larger teams learn to finish episodes sooner, their shorter episodes could reduce their share of collected experience.

To investigate the impact of the sampling strategy on policy performance, we compare three F-MADDPG policies trained using uniform team-size sampling with ordinary replay, inverse-cardinality team-size sampling with ordinary replay, or uniform team-size sampling with bucketed replay. The bucketed baseline adapts ADDRQN's replay strategy by partitioning experience according to initial team size, uniformly selecting a bucket, and then sampling experiences within it. We evaluated all three policies without agent failures and report success rates by initial team size.

\begin{figure}[t]
\centering
\includegraphics[width=\columnwidth]{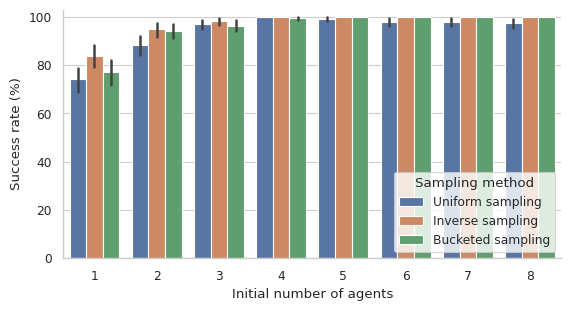}
\caption{Success rates of F-MADDPG policies trained using uniform and inverse-cardinality team-size sampling and bucketed replay sampling. Results are stratified by initial number of agents. Error bars show 95\% confidence intervals.}
\label{fig:sampling-success-rates}
\end{figure}

Figure~\ref{fig:sampling-success-rates} shows the success rates of models trained using the three strategies. Inverse-cardinality sampling achieved the highest or joint-highest success rate at every evaluated team size, increasing the success rate by an average of 3 percentage points over uniform sampling with ordinary replay. The largest difference between these two strategies occurred in the single-agent setting, for which inverse-cardinality sampling improved the success rate by 9.7 percentage points. This improvement is consistent with the sampling rationale. 

Bucketed replay sampling also improved aggregate success relative to uniform team-size sampling with ordinary replay, although the improvement was smaller than that obtained with inverse-cardinality sampling. This experiment compares sampling strategies within F-MADDPG rather than the complete F-MADDPG and ADDRQN algorithms. The relative effectiveness of these strategies may depend on the surrounding architecture and training procedure; the results do not identify which factors account for the observed ranking or establish that inverse-cardinality sampling is generally superior. They support inverse-cardinality sampling as a simple and effective strategy in the evaluated F-MADDPG setting.

\subsection{Robustness to Agent Failures}
\label{sec:failure-exposure}
As discussed in Section~\ref{sec:team-sampling}, reinforcement learning models generally struggle in situations not encountered during training. F-MADDPG is designed to produce policies that are robust to changes in team size. The preceding experiments considered changes between episodes; however, the active team size may also change within an episode when one or more agents terminate.

This section evaluates whether F-MADDPG can maintain task performance after agent failures. We consider two approaches to achieving robustness to such failures. The first is to train the model using a randomly sampled team size in each episode, as described in previous sections. If the model generalizes to every $N_a \in \left[N_a^{\min}, N_a^{\max}\right]$, the loss of one agent reduces the active team from $\lvert\mathcal{I}_t\rvert$ to $\lvert\mathcal{I}_{t+1}\rvert=\lvert\mathcal{I}_t\rvert-1$, a size encountered during training. Failures are disabled when $\lvert\mathcal{I}_t\rvert=1$ because the task cannot be completed without an active agent. The second approach explicitly introduces agent failures during training.

After the active UAVs act in transition $t$, the simulator independently selects each UAV $j\in\mathcal{I}_t$ for failure with probability $p$. The simulator permanently deactivates the selected UAVs while ensuring that at least one UAV remains active.

Tables~\ref{tab:death-success-rate} and~\ref{tab:death-completion-time} report the success rates and mean completion times, respectively, for policies trained and evaluated using three values of $p$, corresponding to no, infrequent, and frequent failures. In both tables, rows indicate the failure probability used during training, whereas columns indicate the probability used during evaluation. Each policy was evaluated under every listed failure probability. For a given training condition, changing the evaluation failure probability had little effect on either metric. Differences among the training conditions were larger but nonmonotonic: training with $p=0.0005$ produced longer completion times, whereas training with $p=0.002$ yielded results closer to those obtained without failures. Thus, the results provide no evidence that greater exposure to failures during training consistently improves policy performance. Results were also similar across initial team sizes; therefore, we do not stratify them by initial team size.

\begin{table}[t]
\centering
\caption{Success rate (\%) under different training and evaluation failure probabilities.}
\label{tab:death-success-rate}
\scriptsize
\setlength{\tabcolsep}{3.5pt}
\begin{tabular}{@{}lrrr@{}}
\toprule
Training $p$ & \multicolumn{3}{c}{Evaluation $p$} \\
\cmidrule(l){2-4}
 & $0$ & $0.0005$ & $0.002$ \\
\midrule
$0$      & 99.60 & 99.60 & 99.66 \\
$0.0005$ & 99.09 & 99.09 & 98.92 \\
$0.002$  & 99.60 & 99.60 & 99.60 \\
\bottomrule
\end{tabular}
\end{table}

\begin{table}[t]
\centering
\caption{Mean completion time (s) under different training and evaluation failure probabilities.}
\label{tab:death-completion-time}
\scriptsize
\setlength{\tabcolsep}{3.5pt}
\begin{tabular}{@{}lrrr@{}}
\toprule
Training $p$ & \multicolumn{3}{c}{Evaluation $p$} \\
\cmidrule(l){2-4}
 & $0$ & $0.0005$ & $0.002$ \\
\midrule
$0$      & 31.22 & 31.26 & 31.16 \\
$0.0005$ & 36.18 & 36.27 & 37.15 \\
$0.002$  & 32.93 & 32.95 & 33.18 \\
\bottomrule
\end{tabular}
\end{table}

\begin{figure}[t]
\centering
\includegraphics[width=\columnwidth]{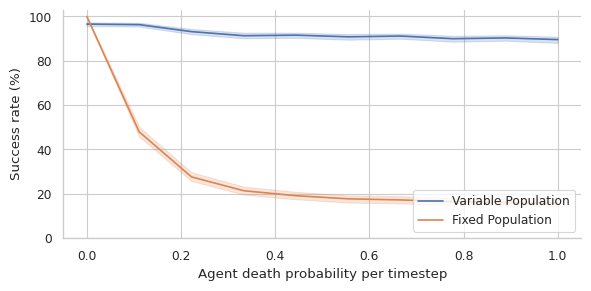}
\caption{Success rates of policies trained with and without population variability, both evaluated with five initial UAVs under different per-step failure probabilities. Shaded areas show 95\% confidence intervals.}
\label{fig:population-variability-ablation}
\end{figure}

These results demonstrate that population variability during training is sufficient for F-MADDPG to learn policies that maintain performance after mid-episode agent failures. To further assess the role of population variability, we compared two policies trained without failures ($p=0$): one trained across different initial team sizes and another trained with a fixed team size of $N_a=5$. Both policies were evaluated with five initial UAVs, varying only the per-agent failure probability per timestep from $p=0$ to $p=1$. The environment prevents the final active UAV from failing. Figure~\ref{fig:population-variability-ablation} shows the resulting success rates. The fixed-population policy's success rate decreases sharply as failure probability increases, whereas the policy trained with variable team sizes maintains a high success rate across the evaluated probabilities.

\subsection{Generalization to Larger Populations}

We evaluate F-MADDPG's ability to generalize beyond the training population range by comparing it with an MLP baseline in scenarios with up to 20 UAVs. The models were trained with inverse-cardinality sampling and no failures on episodes with up to 8 UAVs. These are the same models used in section \ref{sec:training-efficiency}. We evaluated both policies at each fixed team size from 1 to 20 UAVs.

\begin{figure}[t]
\centering
\includegraphics[width=\columnwidth]{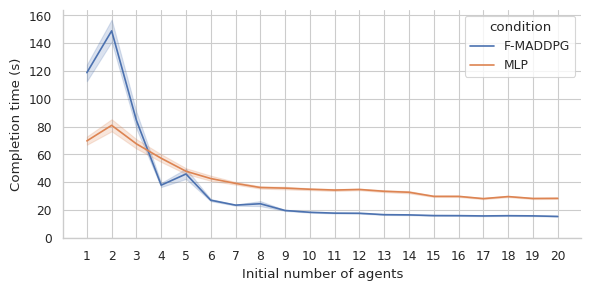}
\caption{Average completion time of F-MADDPG and the MLP baseline for different initial team sizes. Unsuccessful missions are assigned the 500~s episode horizon. Shaded bands show 95\% confidence intervals.}
\label{fig:larger-populations}
\end{figure}

Figure~\ref{fig:larger-populations} shows the average completion time of F-MADDPG and the MLP baseline for different initial team sizes. Mean completion times generally decreased for both policies as team size increased beyond eight UAVs. F-MADDPG achieves lower completion times for larger populations within the training range and retains this advantage beyond the largest training population.

The observation structure described in equation~\ref{eq:observation} limits peer information to a fixed number of closest UAVs. Consequently, the observation structure remains consistent for larger populations. This stability may explain the models' ability to generalize to larger populations. As such, the results may be environment-specific and may not apply to other scenarios where out-of-distribution observations occur as a consequence of population growth. Further experiments are required to evaluate F-MADDPG's generalization capabilities in other environments.

\begin{figure}[t]
\centering
\includegraphics[width=\columnwidth]{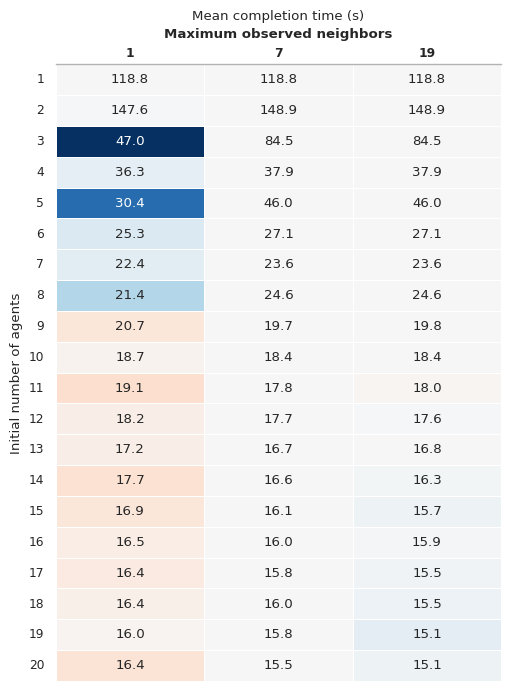}
\caption{Mean completion time of F-MADDPG across population sizes and observation neighborhood limits. Cell values report completion time in seconds. Within each population size, blue indicates shorter completion times and red indicates longer completion times relative to the seven-neighbor variant.}
\label{fig:neighborhood-sizes}
\end{figure}

This observation structure affects our ability to demonstrate one of the benefits of F-MADDPG over the MLP baseline, namely its ability to process variable-cardinality observations. Figure~\ref{fig:neighborhood-sizes} shows the mean completion time of the same trained F-MADDPG policy evaluated with neighborhood limits of 1, 7, and 19 UAVs. Compared with the other two settings, $N_a^*=1$ yielded lower mean completion times for teams of 2--8 UAVs and higher mean completion times for teams of 9--20 UAVs. The $N_a^*=7$ and $N_a^*=19$ settings yielded similar mean completion times. Overall, the results suggest that the benefit of observing more neighbors depends on team size in this environment. Further experiments are required to evaluate F-MADDPG's ability to process variable-cardinality observations in other environments.

\section{Conclusion}

We introduced F-MADDPG, a multi-agent reinforcement learning framework designed for cooperative tasks with variable team sizes and permanent agent losses. Its Flexible Encoder uses masked, field-specific representations to process variable-cardinality observations and centralized critic inputs without treating absent entities as conventional values. Consequently, a single shared policy can operate across different UAV-team and sensor-population sizes and can continue a mission after failed UAVs are removed from the system.

In the evaluated data-harvesting scenarios, F-MADDPG achieved competitive success rates while completing missions faster than the MLP-based baseline. Inverse-cardinality team-size sampling yielded higher aggregate success rates than uniform team-size sampling with either ordinary or bucketed replay, with the largest gains observed for smaller teams. For each trained policy, performance remained similar across the evaluated failure probabilities. Exposure to failures during training yielded no consistent performance gains. This suggests that, in the considered setting, exposure to varying team sizes can be sufficient for learning behavior that remains effective after mid-episode agent terminations, without requiring failures to be explicitly introduced during training. An ablation study supports this hypothesis, as the variable-population policy outperformed a fixed-population baseline in scenarios with increasing probability of agent losses. Moreover, we demonstrate that both F-MADDPG and the MLP baseline generalize well to scenarios with agent populations larger than those observed during training, although the observation structure of the data-harvesting scenario facilitates this.

These findings are limited to a simulated environment with simplified UAV dynamics, globally available team information, and fail-stop failures. Future work should consider communication-constrained observations, more realistic mobility and energy models, and other failure modes, including intermittent communication and degraded rather than complete loss of functionality. Evaluating F-MADDPG in additional cooperative tasks and physical UAV deployments would further clarify the generality and practical robustness of the proposed architecture. The Flexible Encoder can also be extended to represent other input representations, such as spatial or image data.

\bibliographystyle{cas-model2-names}
\bibliography{references}

\end{document}